\documentclass{article}

\usepackage{times}
\usepackage{graphicx} 
\usepackage{subfigure} 

\usepackage{natbib}

\usepackage{algorithm}
\usepackage{algorithmic}

\usepackage{hyperref}

\usepackage{amsmath}

\usepackage[accepted]{icml2017}

\icmltitlerunning{A Deep Learning Approach for Joint Video Frame and Reward Prediction in Atari Games}

\begin{document} 

\twocolumn[
\icmltitle{A Deep Learning Approach for Joint Video Frame and Reward Prediction in Atari Games}



\icmlsetsymbol{equal}{*}

\begin{icmlauthorlist}
\icmlauthor{Felix Leibfried}{prow}
\icmlauthor{Nate Kushman}{msr}
\icmlauthor{Katja Hofmann}{msr}
\end{icmlauthorlist}

\icmlaffiliation{prow}{PROWLER.io, Cambridge, United Kingdom}
\icmlaffiliation{msr}{Microsoft Research, Cambridge, United Kingdom}

\icmlcorrespondingauthor{Felix Leibfried}{felix@prowler.io}

\icmlkeywords{deep learning, model-based learning, reinforcement learning}

\vskip 0.3in
]



\printAffiliationsAndNotice{}  

\begin{abstract} 
Reinforcement learning is concerned with identifying reward-maximizing behaviour policies in environments that are initially unknown. State-of-the-art reinforcement learning approaches, such as deep Q-networks, are model-free and learn to act effectively across a wide range of environments such as Atari games, but require huge amounts of data. Model-based techniques are more data-efficient, but need to acquire explicit knowledge about the environment.

In this paper, we take a step towards using model-based techniques in environments with a high-dimensional visual state space by demonstrating that it is  possible to learn system dynamics and the reward structure jointly. 
Our contribution is to extend a recently developed deep neural network for video frame prediction in Atari games to enable reward prediction as well. To this end, we phrase a joint optimization problem for minimizing both video frame and reward reconstruction loss, and adapt network parameters accordingly.
Empirical evaluations on five Atari games demonstrate accurate cumulative reward prediction of up to 200 frames. We consider these results as opening up important directions for model-based reinforcement learning in complex, initially unknown environments. 
\end{abstract} 

\section{Introduction}
\label{intro}


Reinforcement learning (RL) is concerned with finding optimal behaviour policies to maximize agents' future reward.
Approaches to RL can be divided into model-free and model-based. In \emph{model-free} settings, agents learn by trial and error but do not aim to explicitly capture the dynamics of the environment or the structure of the reward function. State-of-the-art model-free approaches, such as deep Q-networks (DQN, \citep{Mnih2015}), effectively approximate so-called Q-values, i.e. the expected cumulative future reward of taking specific actions in a given state, using deep neural networks. The impressive effectiveness of these approaches comes from their ability to learn complex policies directly from high-dimensional inputs (e.g., video frames). Despite their effectiveness, model-free approaches require substantial amount of training data that has to be collected through direct interactions with the environment, which limits their applicability when sampling such data is expensive (as in most real-world applications). Additionally, model-free RL requires access to reward observations during training, which is problematic in environments with a sparse reward structure---unless coupled with an explicit exploration mechanism.

The second alternative for RL algorithms is \textit{model-based}. Here, agents explicitly gather statistics about the environment or the reward---in a more narrow definition these statistics comprise environment dynamics and the reward function. 
In recent work, model-based techniques were successfully used to learn statistics about cumulative future rewards \citep{Veness2015} and to improve exploration \citep{Pathak2017,Bellemare2016, Oh2015}, resulting in more data efficient learning compared to model-free approaches. 
When an accurate model of the true environment dynamics and the true reward function is available, model-based approaches such as planning via Monte-Carlo tree search \citep{Browne2012} outperform model-free state-of-the-art approaches \citep{Guo2014}.

A key open question then is whether effective model-based RL is possible in complex settings where the environment dynamics and the reward function are both initially unknown, and the agent has to acquire such knowledge through experience. In this paper, we take a step towards addressing this question by extending recent work on video frame prediction \citep{Oh2015}, which effectively learns system dynamics in the Arcade Learning Environment for Atari games (ALE, \citep{Bellemare2013}),
to enable reward prediction as well. The resulting approach is a deep convolutional neural network that enables joint prediction of future states and rewards using a single latent representation. Network parameters are trained by minimizing a joint optimization objective minimizing both video frame and reward reconstruction loss. Such a joint prediction model is a necessary prerequisite for model-based algorithms such as Dyna learning \citep{Sutton1990} or planning with Monte-Carlo tree search \citep{Browne2012}.

We evaluate our approach on five Atari games.
Our empirical results demonstrate successful prediction of cumulative rewards up to roughly 200 frames. We complement our quantitative results with a qualitative error analysis by visualizing example predictions. Our results are the first to demonstrate the feasibility of using a learned dynamics and reward model for model-based RL.

The rest of our paper is structured as follows. In Section~\ref{mot}, we discuss related work and motivate our approach further. In Section~\ref{net}, we describe the network architecture and the training procedure. In Section~\ref{res}, we present our results on cumulative reward prediction. In Section~\ref{con}, we conclude and outline future work.

\section{Related Work and Motivation}
\label{mot}

Two lines of research are related to the work presented in this paper: model-based RL and optimal control theory. Model-based RL utilizes a given or learned model of some aspect of a task to, e.g., reduce data or exploration requirements \citep{Bellemare2016,Oh2015,Veness2015}. Optimal control theory describes mathematical principles for deriving control policies in continuous action spaces that maximize cumulative future reward in scenarios with known system dynamics and known reward structure \citep{Bertsekas2007,Bertsekas2005}.

There has been recent interest in combining principles from optimal control theory and model-based learning in settings where no information on system dynamics is available a priori and instead has to be acquired from visual data \citep{Finn2016,Wahlstrom2015,Watter2015}. The general idea behind these approaches is to learn a compressed latent representation of the visual state space from raw images through autoencoder networks \citep{Bengio2009} and to utilize the acquired latent representation to infer system dynamics. System dynamics are then used to specify a planning problem which can be solved by optimization techniques to derive optimal policies. 
\cite{Watter2015} introduce an approach for learning system dynamics from raw visual data by jointly training a variational autoencoder \citep{Kingma2014,Rezende2014} and a state prediction model 
that operates in the autoencoder's compressed latent state representation. A similar approach for jointly learning a compressed state representation and a predictive model is pursued by \cite{Wahlstrom2015}.
 \cite{Finn2016} devise a sequential approach that first learns a latent state representation from visual data 
and that subsequently exploits this latent representation to augment a robot's initial state space describing joint angles and end-effector positions. The augmented state space is used to improve estimates of local dynamics for planning.

The approaches presented above assume knowledge of the functional form of the true reward signal and are hence not directly applicable in settings like ALE (and many real-world settings) where the reward function is initially unknown. Planning in such settings therefore necessitates learning both system dynamics and the reward function in order to infer optimal behavioral policies. Recent work by \cite{Oh2015} introduced an approach for learning environment dynamics from pixel images and demonstrated that this enabled successful video frame prediction over up to 400 frames. In our current paper, we extend this recent work to enable reward prediction as well by modifying the network's architecture and training objective accordingly. The modification of the training objective bears a positive side effect: since our network must optimize a compound loss consisting of the video frame reconstruction loss and the reward loss, reward-relevant aspects in the video frames to which the reconstruction loss alone might be insensitive are explicitly captured by the optimization objective. In the subsequent section, we elucidate the approach from \cite{Oh2015} as well as our extensions in more detail.

\section{Network Architecture and Training}
\label{net}

The deep network proposed by \cite{Oh2015} for video frame prediction in Atari games 
aims at learning a function that  predicts the video frame $\mathbf{s}_{t+1}$ at the next time step $t+1$, given the current history of frames $\mathbf{s}_{t-h+1:t}$ with time horizon $h$ and the current action $\mathbf{a}_t$ taken by the agent---see Section~\ref{vid_pred}. We extend this work to enable joint video frame and reward prediction such that the network anticipates the current reward $\mathbf{r}_t$---see Sections~\ref{rew_pred} and~\ref{train}.

\subsection{Video Frame Prediction}
\label{vid_pred}

\begin{figure*}[t!]
\begin{center}
\includegraphics[width=1\linewidth]{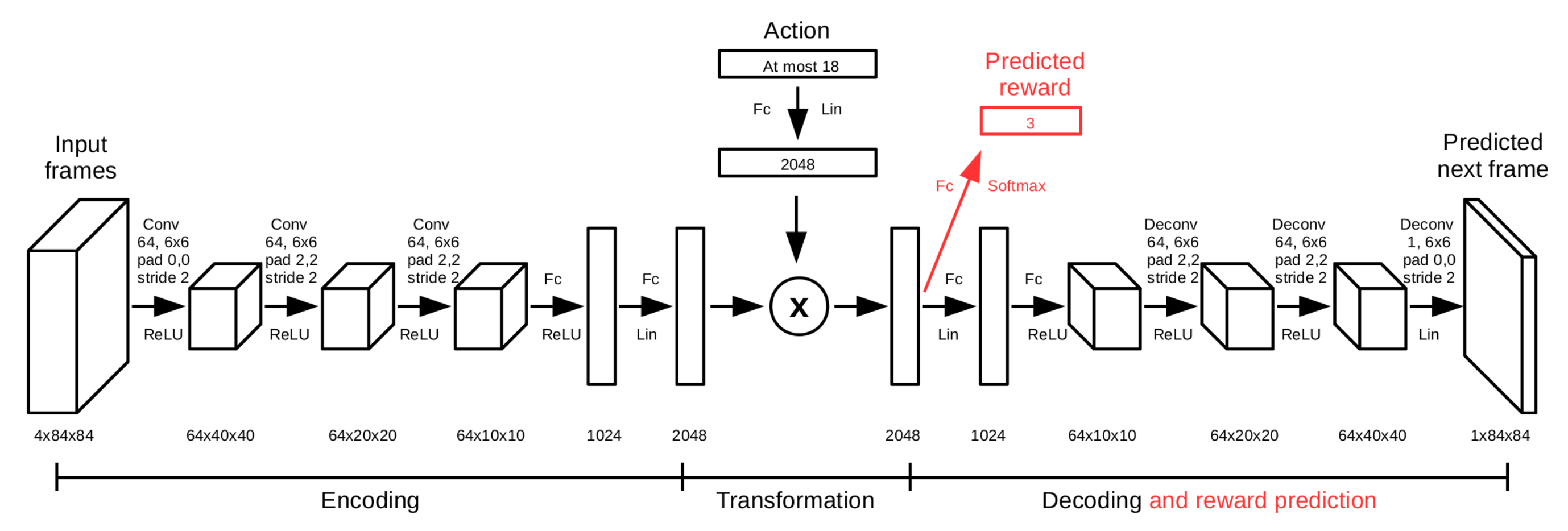}
\caption{Network architecture for joint video frame and reward prediction. The architecture comprises three stages: an encoding stage mapping current input frames to some compressed latent representation, a transformation stage integrating the current action into the latent representation through element-wise vector multiplication denoted by '$\times$', and a final predictive stage for reconstructing the frame of the next time step and the current reward. The network uses three different types of neuron layers ('Conv' for  convolutional, 'Deconv' for deconvolutional and 'Fc' for forward connection) in combination with three different types of activation functions ('ReLU', 'Softmax' and 'Lin'  for linear activations). The dimensional extend of individual layers is either depicted beneath or within layers. The network part coloured in red highlights the extension for reward prediction.}
\label{fig:model}
\end{center}
\end{figure*}
The video-frame-predictive architecture from \cite{Oh2015} comprises three information-processing stages: an encoding stage that maps input frames to some compressed latent representation, a transformation stage that integrates the current action into the compressed latent representation, and a decoding stage that maps the compressed latent representation to the predicted next frame---see Figure~\ref{fig:model}. The initial encoding stage is a sequence of convolutional and forward operations that map the current frame history $\mathbf{s}_{t-h+1:t}$---a three-dimensional tensor---to a compressed feature vector $\mathbf{h}^{\text{enc}}_t$. The transformation stage converts this compressed feature vector $\mathbf{h}^{\text{enc}}_t$ into an action-conditional representation $\mathbf{h}^{\text{dec}}_t$ in vectorized form by integrating the current action $\mathbf{a}_t$. The current action $\mathbf{a}_t$ is represented as a one-hot vector with length varying from game to game since there are at least 3 and at most 18 actions in ALE. The integration of the current action into the compressed feature vector includes an element-wise vector multiplication---depicted as '$\times$' in Figure~\ref{fig:model}---with the particularity that the two neuron layers involved in this element-wise multiplication are the only layers in the entire network without bias parameters, see Section 3.2 in \cite{Oh2015}. Finally, the decoding stage performs a series of forward and deconvolutional operations \citep{Dosovitskiy2015,Zeiler2010} by mapping the action-conditional representation $\mathbf{h}^{\text{dec}}_t$ of the current frame history $\mathbf{s}_{t-h+1:t}$ and the current action $\mathbf{a}_t$ to the predicted video frame $\mathbf{s}_{t+1}$ of the next time step $t+1$. Note that this necessitates a reshape operation at the beginning of the decoding cascade in order to transform the vectorized hidden representation into a three-dimensional tensor.
The whole network uses linear and rectified linear units \citep{Glorot2011} only. In all our experiments, following DQN \citep{Mnih2015}, the video frames processed by the network are $84 \times 84$ grey-scale images down-sampled from the full-resolution $210 \times 160$ Atari RGB images from ALE. Following \cite{Mnih2015} and \cite{Oh2015}, the history frame time horizon $h$ is set to $4$.

\subsection{Reward Prediction}
\label{rew_pred}

In this section we detail our proposed network architecture for joint video frame and reward prediction.  
Our model assumes ternary rewards which result from reward clipping in line with \cite{Mnih2015}. Original game scores in ALE are integers that can vary significantly between different Atari games and the corresponding original rewards are clipped to assume one of three values: $-1$ for negative rewards, $0$ for no reward and $1$ for positive rewards. Because of reward clipping, rewards can be represented as vectors $\mathbf{r}_t$ in one-hot encoding of size 3. 

In Figure~\ref{fig:model}, our extension of the video-frame-predictive architecture from \cite{Oh2015} to enable reward prediction is highlighted in red. We add an additional softmax layer to predict the current reward $\mathbf{r}_t$ with information contained in the action-conditional encoding $\mathbf{h}^{\text{dec}}_t$. The motivation behind this extension is twofold. First, our extension makes it possible to jointly train the network with a compound objective that emphasizes both video frame reconstruction and reward prediction, and thus encourages the network to not abstract away reward-relevant features to which the reconstruction loss alone might be insensitive.
Second, this formulation facilitates the future use of the model 
for reward prediction through virtual roll-outs in the compressed latent space, without the computational expensive necessity of reconstructing video frames explicitly---note that this requires another "shortcut" predictive model to map from $\mathbf{h}^{\text{dec}}_t$ to $\mathbf{h}^{\text{enc}}_{t+1}$.

Following previous work \citep{Oh2015,Mnih2015}, actions are chosen by the agent on every fourth frame and are repeated on frames that were skipped. Skipped frames and repeated actions are hence not part of the data sets used to train and test the predictive network on, and original reward values are accumulated over four frames before clipping. 

\subsection{Training}
\label{train}

Training the model for joint video frame and reward prediction requires trajectory samples $\left\{ \left( \mathbf{s}_n^{(i)}, \mathbf{a}_n^{(i)}, \mathbf{r}_n^{(i)} \right)_{n=1}^N \right\}_{i=1}^I$ collected by some agent playing the Atari game, where  
$i$ is an index over trajectories and $n$ is a time index over samples within one trajectory $i$. The parameter $I$ denotes the number of trajectories in the training set or the minibatch respectively and the parameter $N$ denotes the length of an individual trajectory. In our case, we trained DQN-agents according to \cite{Mnih2015} and collected trajectory samples after training had finished.

The original training objective in \cite{Oh2015} consists of a video frame reconstruction loss in terms of a squared loss function aimed at minimizing the square of the $l^2$-norm of the difference vector between the ground truth image and its action-conditional reconstruction. We extend this training objective to enable joint reward prediction. This results in a compound training loss consisting of the original video frame reconstruction loss and a reward prediction loss---given by the cross entropy loss that has been proven of value in classification problems \citep{Simard2003}---between the ground truth reward and the predicted reward:
\begin{equation*}
\begin{split}
L_K(\theta) & = \frac{1}{2 \cdot I \cdot T \cdot K} \sum_{i=1}^I \sum_{t=0}^{T-1} \sum_{k=1}^K \left( \underbrace{ \left| \left| \mathbf{s}_{t+k}^{(i)} - \hat{\mathbf{s}}_{t+k}^{(i)} \right| \right|_2^2 }_{\text{video frame reconstruction loss}} \right.  \\
& \left. + \lambda \cdot \underbrace{ (-1)  \sum_{l=1}^3 \mathbf{r}_{t+k}^{(i)}[l] \cdot \ln \mathbf{p}_{t+k}^{(i)}[l] }_{\text{reward prediction loss}} \right)
\end{split}
\end{equation*}
where $\hat{\mathbf{s}}_{t+k}^{(i)}$ denotes the $k$-step look ahead frame prediction with target video frame $\mathbf{s}_{t+k}^{(i)}$ and $\mathbf{p}_{t+k}^{(i)}$ denotes the $k$-step look ahead probability values of the reward-predicting softmax layer---depicted in red in Figure\ref{fig:model}---with target reward vector $\mathbf{r}_{t+k}^{(i)}$. Note that reward values are ternary because of reward clipping. The parameter $\lambda>0$ controls the trade-off between video frame reconstruction and reward loss. The parameter $T$ is a time horizon parameter that determines how often a single trajectory sample $i$ is unrolled into the future, and $K$ determines the look ahead prediction horizon dictating how far the network predicts into the future by using its own video frame predicted output as input for the next time step. 

Following \cite{Oh2015} and \cite{Michalski2014}, we apply a curriculum learning \citep{Bengio2009a} scheme by successively increasing $K$ in the course of training such that the network initially learns to predict over a short time horizon and becomes fine-tuned on longer-term predictions as training advances. For training details and hyperparameter settings, see Sections~\ref{train_det}, \ref{rew_weight_exp}, \ref{grad_clip_exp} and~\ref{cur_learn_exp}. The network parameters $\theta$ are updated by stochastic gradient descent, derivatives of the training objective are computed with backpropagation through time \citep{Werbos1988}.

\section{Results}
\label{res}

In our evaluations, we investigate cumulative reward predictions quantitatively and qualitatively on five different Atari games (Q*bert, Seaquest, Freeway, Ms Pacman and Space Invaders). 
The quantitative analysis comprises evaluating the cumulative reward prediction error---see Section~\ref{cumulative_rew_pred}. The qualitative analysis comprises visualizations of example predictions in Seaquest---see Section~\ref{traj_sea}. 

\subsection{Quantitative Reward Prediction Analysis: Cumulative Reward Error}
\label{cumulative_rew_pred}

Our quantitative evaluation examines whether our joint model of system dynamics and reward function results in a shared latent representation that enables accurate cumulative reward prediction. We assess cumulative reward prediction on test sets consisting of approximately $50$,$000$ video frames per game, including actions and rewards. Each network is evaluated on $1$,$000$ trajectories---suitable to analyze up to $100$-step ahead prediction---drawn randomly from the test set. Look ahead prediction is measured in terms of the cumulative reward error which is the difference between ground truth cumulative reward and predicted cumulative reward. For each game, this results in $100$ empirical distributions over the cumulative reward error---one distribution for each look ahead step---consisting of $1$,$000$ samples each (one for each trajectory). We compare our model predictions to a baseline model that samples rewards from the marginal reward distribution observed on the test set for each game. 
Note that negative reward values are absent in the games investigated here.

\begin{figure*}[!htb]
\begin{center}
\includegraphics[width=0.9\linewidth]{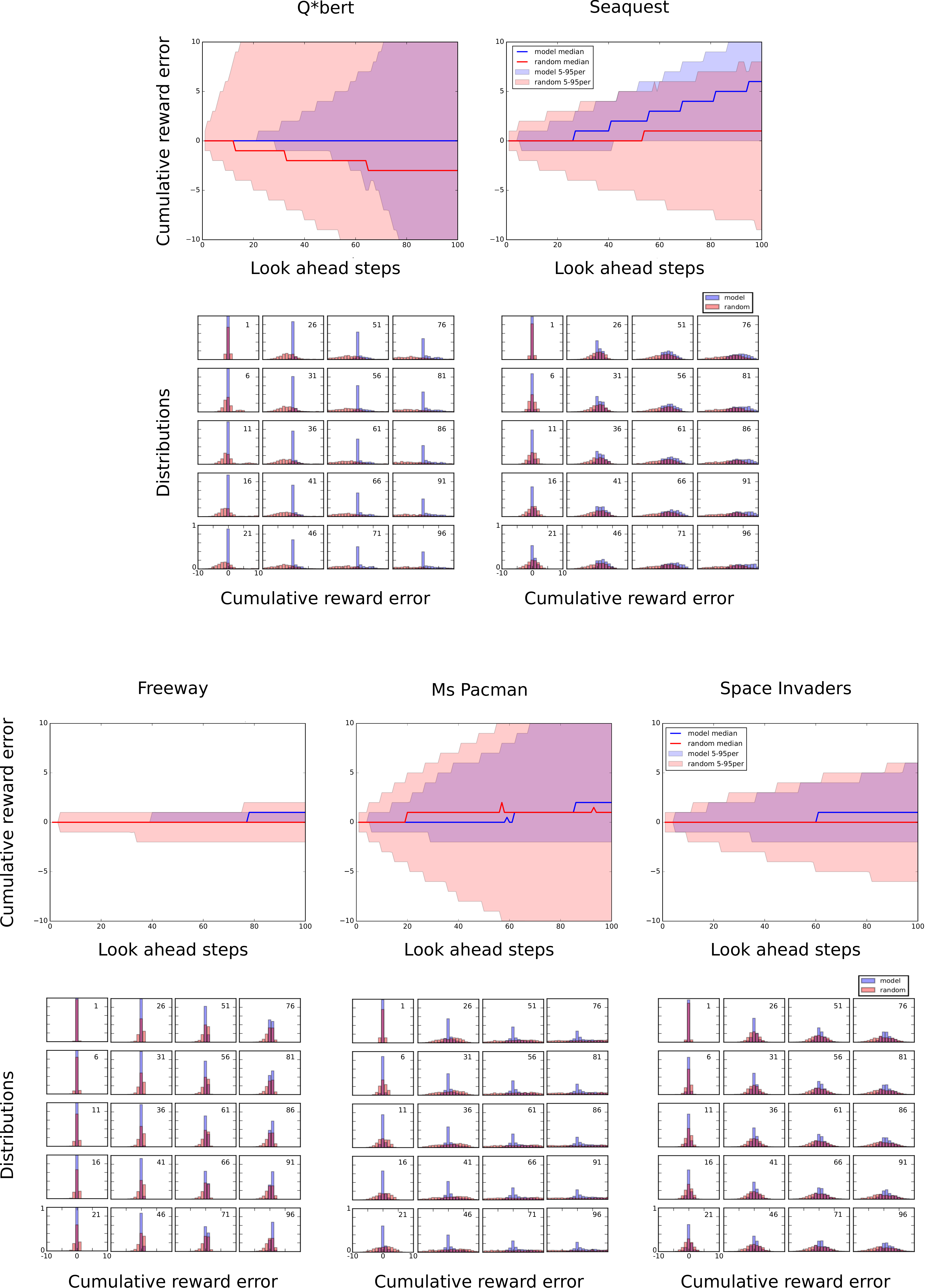}
\caption{Cumulative reward error over look ahead steps in five different Atari games. There are two plots for each game. The top plot per game shows how the median and the 5~to 95~percentiles of the cumulative reward error evolve over look ahead steps for both our model (in blue) and a baseline model that samples rewards from the marginal reward distribution of the test set (in red). Each vertical slice of this concise representation corresponds to a single empirical distribution over the cumulative reward error. We depict these for every fifth look ahead step in the compound plots below for both models. 
These empirical error distributions demonstrate successful cumulative reward prediction over at least 20 steps (80 frames) in all five games as evidenced by their zero-centered and unimodal shape in the first column of each compound plot per game.}
\label{fig:cumul_err}
\end{center}
\end{figure*}

Figure~\ref{fig:cumul_err} illustrates $20$ of the $100$ empirical cumulative reward error distributions in all games for our network model in blue and for the baseline model in red (histograms, bottom), together with the median and the 5~to 95~percentiles of the cumulative reward error over look ahead steps (top).
Across all games, we observe that our joint video frame and reward prediction model accurately predicts future cumulative rewards at least $20$ look ahead steps, and that it predicts future rewards substantially more accurately than the baseline model. 
This is evidenced by cumulative reward error distributions that maintain a unimodal form with mode zero and do not flatten out as quickly as the distributions for the random-prediction baseline model. 
Best results are achieved in Freeway and Q*bert where the probability of zero cumulative reward error at 51 look ahead steps is still around $80\%$ and $60\%$ respectively---see Figure~\ref{fig:cumul_err}. Note that 51 look ahead steps correspond to 204 frames because the underlying DQN agent, collecting trajectory samples for training and testing our model, skipped every fourth frame when choosing an action---see Section~\ref{rew_pred}. 
Lowest performance is obtained in Seaquest where the probability of zero cumulative reward error at 26 steps (104 frames) is around $40\%$ and begins to flatten out soon thereafter---see Figure~\ref{fig:cumul_err}. Running the ALE emulator at a frequency of 60fps, 26 steps correspond to more than 1 second real-time game play because of frame skipping. Since our model is capable of predicting 26 steps ahead in less than 1 second, our model enables real-time planning and could be therefore utilized in an online fashion.

We now turn our attention to error analysis. While the look ahead step at which errors become prominent differs substantially from game to game, we find that overall our model underestimates cumulative reward. This can be seen in the asymmetry towards positive cumulative reward error values when inspecting the 5~to 95~percentile intervals in the first plot per each game in Figure~\ref{fig:cumul_err}. 
We identify a likely cause in (pseudo-)stochastic transitions inherent in these games. Considering Seaquest as our running example, objects such as divers and submarines 
can enter the scene randomly from the right and from the left and at the same time have an essential impact on which rewards the agent can potentially collect. In the ground truth trajectories, the agent's actions are reactions to these objects. If the predicted future trajectory deviates from the ground truth, targeted actions such as shooting will miss their target, leading to underestimating true reward. 
We analyze this effect in more detail in 
Section~\ref{traj_sea}. 

All our experiments were conducted in triplicate with different initial random seeds. Different initial random seeds did not have a significant impact on cumulative reward prediction in all games except Freeway---see Section~\ref{eff_seed_exp} for a detailed analysis. So far, we discussed results concerning reward prediction only. In the appendix, we also evaluate the joint performance of reward and video frame prediction on the test set in terms of the optimization objective as in \cite{Oh2015}, where the authors report successful video frame reconstruction up to approximately 100 steps (400 frames), and observe similar results---see Section~\ref{test_loss}. 

Finally, we could also ascertain that using a \textit{joint training objective} for joint video frame and reward prediction is beneficial by comparing to alternative training methods with separate objectives for video frame reconstruction and reward prediction. To this end, we conducted additional experiments with two baseline models. The first baseline model uses the same network architecture as in Figure~\ref{fig:model} but with a \textit{decoupled training objective} where the reward prediction part is trained using the latent representation of the video frame prediction part as input---this means that gradient updates with respect to reward prediction do not impact the parameters for video frame prediction. The second baseline model uses a \textit{decoupled architecture} with two completely separate convolutional networks---one for video frame prediction and one for reward prediction---without shared parameters. The overall results of these additional experiments are that the joint training objective and the decoupled architecture work better than the decoupled training objective. The added benefit of the joint training objective over the decoupled architecture is a significant reduction in the overall number of parameters due to the shared network layers, see Section~\ref{comparison} for details.

\subsection{Qualitative Reward Prediction Analysis: Example Predictions in Seaquest}
\label{traj_sea}

\begin{figure*}[h]
\begin{center}
\includegraphics[width=0.9\linewidth]{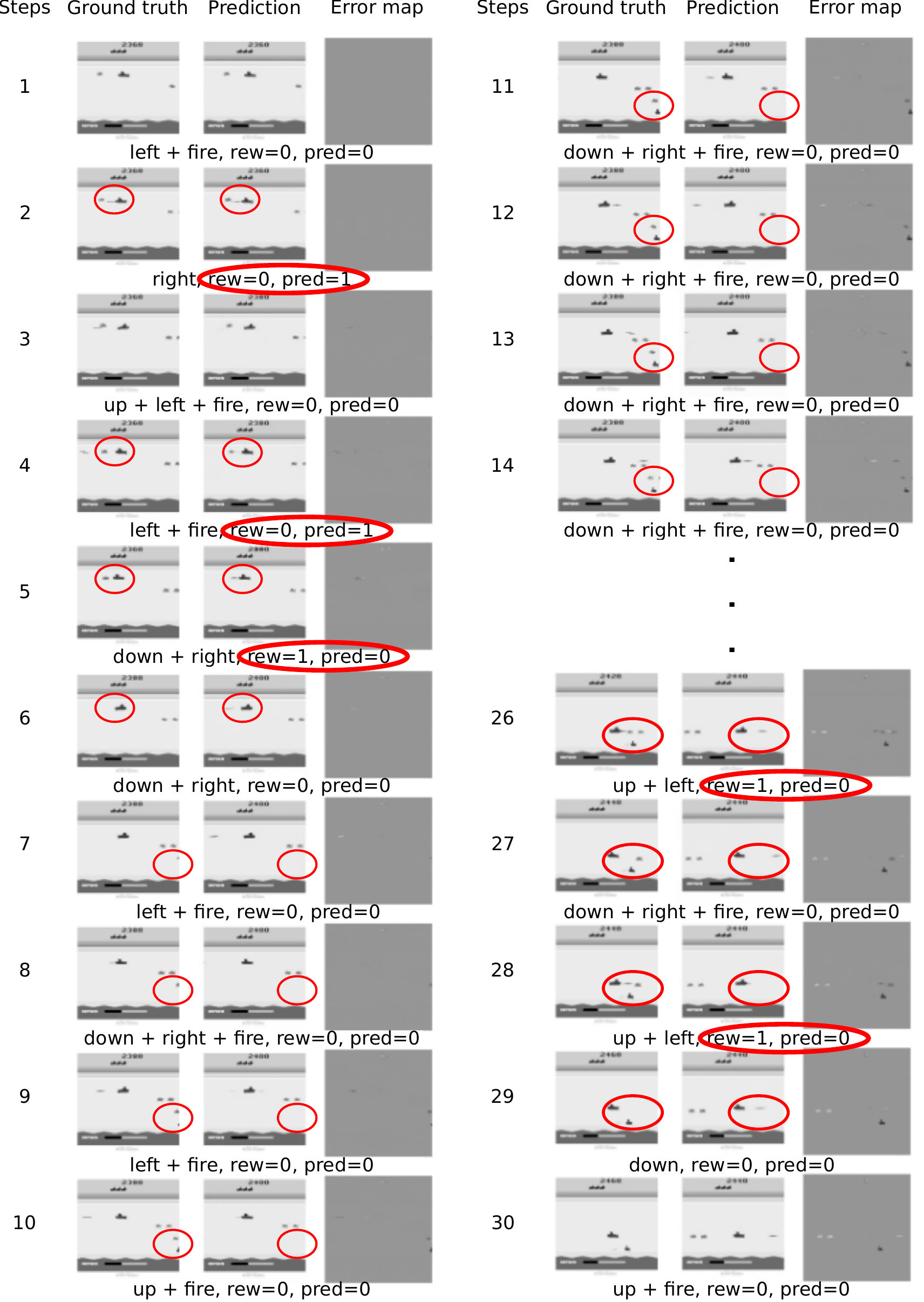}
\caption{Example predictions in Seaquest. Ground truth video frames, model predictions and error maps emphasizing differences between ground truth and predicted frames---in form of the squared error between pixel values---are compared column-wise. Error maps highlight objects in black or white respectively depending on whether these objects are absent by mistake or present by mistake in the model's prediction. Actions taken by the agent as well as ground truth rewards ('rew') and reward predictions ('pred') are shown below video and error frames. Peculiarities in the prediction process are marked in red. The figure demonstrates how our predictive model fails to anticipate objects that randomly enter the scene from the right and rewards associated to these objects.}
\label{fig:trajecs_s}
\end{center}
\end{figure*}

In the previous section, we identified stochasticity in state transitions as a likely cause for relatively low performance in long-term cumulative reward prediction in games such as Seaquest. In Seaquest objects may randomly enter a scene in a non-deterministic fashion. Errors in predicting these events result in predicted possible futures that do not match actually observed future states, resulting in inaccurate reward predictions. 
Here, we support this hypothesis by visualizations in Seaquest illustrating joint video frame and reward prediction for a single network over 20 steps (80 frames)---see Figure~\ref{fig:trajecs_s} where ground truth video frames are compared to predicted video frames in terms of error maps. Error maps emphasize the difference between ground truth and predicted frames through squared error values between pixels in black or white depending on whether objects are absent or present by mistake in the network's prediction. Actions, ground truth rewards and model-predicted rewards are shown between state transitions. Peculiarities in the prediction are shown in red. 

In step 2, the model predicts reward by mistake because the agent barely misses its target. Steps 4 to 6 report how the model predicts reward correctly but is off by one time step. Steps 7 to 14 depict problems caused by objects randomly entering the scene from the right which the model cannot predict. Steps 26 to 30 show problems to predict rewards at steps 26 and 28 as these rewards are attached to objects the model failed to notice entering the scene earlier. 

\section{Conclusion and Future Work}
\label{con}

In this paper, we extended recent work on video frame prediction \citep{Oh2015} in Atari games to enable reward prediction. Our approach can be used to jointly predict video frames and cumulative rewards up to a horizon of approximately 200 frames in five different games (Q*bert, Seaquest, Freeway, Ms Pacman and Space Invaders). We achieved best results in Freeway and Q*bert where the probability of zero cumulative reward error after 200 frames is still around $80\%$ and $60\%$ respectively, and worst results in Seaquest where the probability of zero cumulative reward error after 100 frames is around $40\%$. We compared our model to a random prediction baseline model as well as to more sophisticated models with separate objectives for video frame and reward prediction, ascertaining the benefits of a joint training objective.

Our study fits into the general line of research using autoencoder networks to learn a latent representation from visual data \citep{Finn2016, Goroshin2015, Gregor2015, Kulkarni2015, Srivastava2015, Wahlstrom2015, Watter2015, Kingma2014, Rezende2014, Lange2012, Hinton2011, Ranzato2007}, and extends this line of research by showing that autoencoder networks are capable of learning a combined representation for system dynamics and the reward function in reinforcement learning settings with high-dimensional visual state spaces---a first step towards applying model-based techniques (such as planning for example) in environments where the reward function is not initially known. 

Our positive results open up intriguing directions for future work. Our long-term goal is the integration of model-based and model-free approaches for effective interactive learning and planning in complex environments. Directions for achieving this long-standing challenge include the Dyna method \citep{Sutton1990}, which uses a predictive model to artificially augment expensive training data, and has been shown to lead to substantial reductions in data requirements in tabular RL approaches. Recently a Dyna-style algorithm was   combined with DQN in environments with a continuous action space and a known reward function \citep{Gu2016} as opposed to ALE where actions are discrete and the Atari game's reward function is unknown.

Alternatively, the model could be utilized for planning via Monte-Carlo tree search \citep{Guo2014,Browne2012}. Recently, a similar approach as we proposed in this work was used to jointly learn system dynamics and the reward function in the Atari domain and applied in the context of planning with Monte-Carlo tree search \citep{Fu2016}. This procedure could not compete with state-of-the-art results achieved by the model-free DQN baseline due to compounding errors in the predictions. We did not observe compounding errors in our approach, instead we demonstrated reliable joint video frame and reward prediction up to 200 frames establishing hence a necessary prerequisite for planning in high-dimensional environments like Atari games with unknown dynamics and reward function. 

We hypothesize that model-based reinforcement learning agents using our proposed model for joint video frame and reward prediction presented in this work are particularly beneficial in multi-task or life-long learning scenarios where the reward function changes but the environment dynamics are stationary. Testing this hypothesis requires a flexible learning framework where the reward function and the artificial environment can be changed by the experimenter in an arbitrary fashion, which is not possible in ALE where the environment and the reward function are fixed per game. A learning environment providing such a flexibility  is the recently released Malm\"o platform for Minecraft \citep{Johnson2016} where researchers can create user-defined environments and tasks in order to evaluate the performance of artificial agents.

Joint video frame and reward prediction should also improve upon exploration in high-dimensional environments with sparse reward signals. In the seminal work of \cite{Oh2015}, the authors could demonstrate improved exploration in some Atari games by encouraging the agent to visit novel predicted states that are dissimilar to states visited before. In \cite{Pathak2017}, the authors propose a model-based exploration scheme that triggers an intrinsic reward signal whenever the agent enters a state with high prediction error according to a state prediction model yielding impressive exploration behaviour in environments like VizDoom and Super Mario Bros. Importantly, both aforementioned exploration schemes do not include a reward prediction model and could be improved further upon by encouraging the agent to visit novel states that are potentially reward-yielding or with high reward prediction error.

This paper aims to establish a necessary prerequisite for model-based learning in environments with unknown dynamics and an unknown reward function, where both need to be learnt from visual input. Model-based techniques have been applied in the past in scenarios with non-visual state spaces where the reward function is known but the system dynamics are not, and where system dynamics are approximated via locally linear Gaussian assumptions \cite{Gu2016, Deisenroth2011}. While these approaches are sample-efficient, they are unsuited for high-dimensional state representations.

Extensions of our model to non-deterministic state transitions via dropout \citep{Srivastava2014} and variational autoencoders \citep{Kingma2014,Rezende2014} is a promising direction to alleviate the limitations provoked by (pseudo-)random events. Pseudo-random events are the most likely cause for biased reward prediction which was most severe in Seaquest, see Section~\ref{traj_sea}. Predicting alternative versions of the future could address the bias in the reward prediction process.
Our work is similar to recent studies in model-based learning \citep{Pascanu2017,Wang2017,Weber2017}. The advantage of our approach is to potentially enable efficient planning in low-dimensional latent spaces when the observed state space is high-dimensional. 

\clearpage
\bibliography{ICLR_paper}
\bibliographystyle{icml2017}

\appendix
\section{Appendix}

\subsection{Training Details}
\label{train_det}

We performed all our experiments in Python with Chainer and adhered to the instructions in \cite{Oh2015} as close as possible. Trajectory samples for learning the network parameters were obtained from a previously trained DQN agent according to \cite{Mnih2015}. The dataset for training comprised around $500,000$ video frames per game in addition to actions chosen by the DQN agent and rewards collected during game play. Video frames used as network input were $84 \times 84$ grey-scale images with pixel values between 0 and 255 down-sampled from the full-resolution $210 \times 160$ ALE RGB images. We applied a further preprocessing step by dividing each pixel by 255 and subtracting mean pixel values from each image leading to final pixel values $\in [-1;1]$. 

A detailed network architecture is shown in Figure~\ref{fig:model} in the main paper. All weights in the network were initialized according to \cite{Glorot2010} except for those two layers that participate in the element-wise multiplication in Figure~\ref{fig:model}: the weights of the action-processing layer were initialized uniformly in the range $[-0.1;0.1]$ and the weights of the layer receiving the latent encoding of the input video frames were initialized uniformly in the range $[-1;1]$. Training was performed for $1,500,000$ minibatch iterations with a curriculum learning scheme increasing the look ahead parameter $K$ every $500,000$ iterations from 1 to 3 to 5. When increasing the look ahead parameter $K$ for the first time after $500,000$ iterations, the minibatch size $I$  was also altered from 32 to 8 as was the learning rate for parameter updates from $10^{-4}$ to $10^{-5}$. Throughout the entire curriculum scheme, the time horizon parameter determining the number of times a single trajectory is unrolled into the future was $T=4$. The optimizer for updating weights was Adam \citep{Kingma2015} with gradient momentum $0.9$, squared gradient momentum $0.95$ and epsilon parameter $10^{-8}$. In evaluation mode, network outputs were clipped to $[-1;1]$ so that strong activations could not accumulate over roll-out time in the network.

In our experiments, we modified the reward prediction loss slightly in order to prevent exploding gradient values by replacing the term $-\ln p$ with a first-order Taylor approximation for $p$-values smaller than $e^{-10}$---a similar technique is used in DQN \citep{Mnih2015} to improve the stability of the optimization algorithm. To identify optimal values for the reward weight $\lambda$, we performed initial experiments on Ms Pacman without applying the aforementioned curriculum learning scheme instead using a fixed look ahead parameter $K=1$. We evaluated the effect of different $\lambda$-values $\in \{ 0.1,1,10,100 \}$ on the training objective and identified $\lambda=1$ for conducting further experiments---see Section~\ref{rew_weight_exp}. After identifying an optimal reward weight, we conducted additional initial experiments without curriculum learning with fixed look ahead parameter $K=1$ on all of the five different Atari games used in this paper. We observed periodic oscillations in the reward prediction loss of the training objective in Seaquest, which was fixed by adding gradient clipping \citep{Pascanu2013} with threshold parameter 1 to our optimization procedure---experiments investigating the effect of gradient clipping in Seaquest are reported in Section~\ref{grad_clip_exp}. The fine-tuning effect of curriculum learning on the training objective in our final experiments is shown in Section~\ref{cur_learn_exp} for all of the five analysed Atari games.

\subsection{Effect of Reward Weight in Ms Pacman}
\label{rew_weight_exp}

\begin{figure*}[h]
\begin{center}
\includegraphics[width=0.8\linewidth]{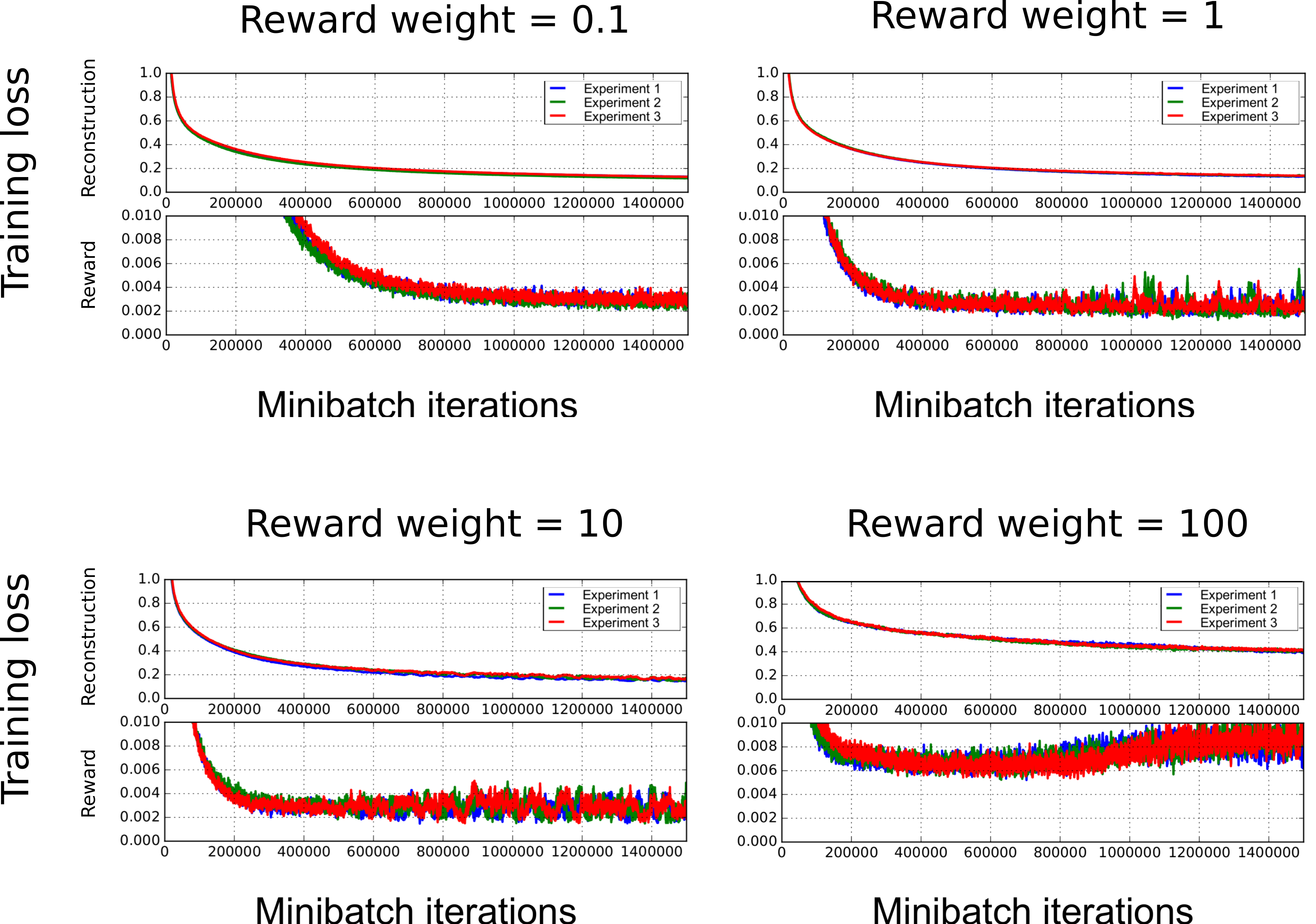}
\caption{Effect of reward weight on training loss in Ms Pacman. Each of the four panels depicts one experiment with a different reward weight $\lambda$. Each panel shows how the training loss evolves over minibatch iterations in terms of two subplots reporting video frame reconstruction and reward loss respectively. Each experiment was conducted three times with different initial random seeds depicted in blue, green and red. Graphs were smoothed with an exponential window of size 1000.}
\label{fig:eff_rew_weight}
\end{center}
\end{figure*}

To identify optimal values for the reward weight $\lambda$, we conducted initial experiments in Ms Pacman without curriculum learning and a fixed look ahead horizon $K=1$. We tested four different $\lambda$-values $\in \{ 0.1, 1, 10, 100\}$ and investigated how the frame reconstruction loss and the reward loss of the training objective evolve over minibatch iterations---see Figure~\ref{fig:eff_rew_weight}. Best results were obtained for $\lambda = 1$ and for $\lambda =10$, whereas values of $\lambda=0.1$ and $\lambda=100$ lead to significantly slower convergence and worse overall training performance respectively.

\subsection{Effect of Gradient Clipping in Seaquest}
\label{grad_clip_exp}

\begin{figure*}[h]
\begin{center}
\includegraphics[width=1\linewidth]{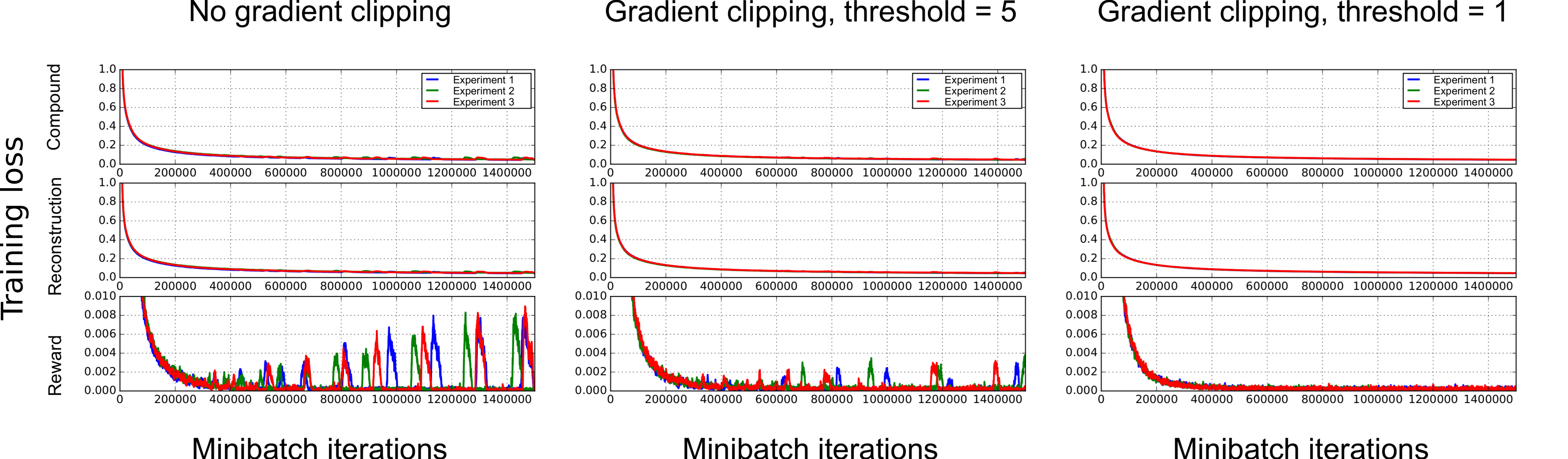}
\caption{Effect of gradient clipping on training loss in Seaquest. The three panels compare experiments with no reward clipping to those with reward clipping using the threshold values 5 and 1 respectively. Subplots within each panel are similar to those in Figure~\ref{fig:eff_rew_weight} but display in the first row the evolution of the compound training loss in addition to the frame reconstruction and reward loss.}
\label{fig:eff_grad_clip}
\end{center}
\end{figure*}

After identifying an optimal value for the reward weight, see Section~\ref{rew_weight_exp}, we observed oscillations in the reward loss of the training objective in Seaquest---see first column in Figure~\ref{fig:eff_grad_clip}---which was solved by adding gradient clipping to our optimization procedure---see second and third column in Figure~\ref{fig:eff_grad_clip}. We tested two different values for the gradient clipping threshold (5 and 1) both of which worked, but for a value of 1 the oscillation vanished completely.

\subsection{Effect of Curriculum Learning}
\label{cur_learn_exp}

\begin{figure*}[h]
\begin{center}
\includegraphics[width=1\linewidth]{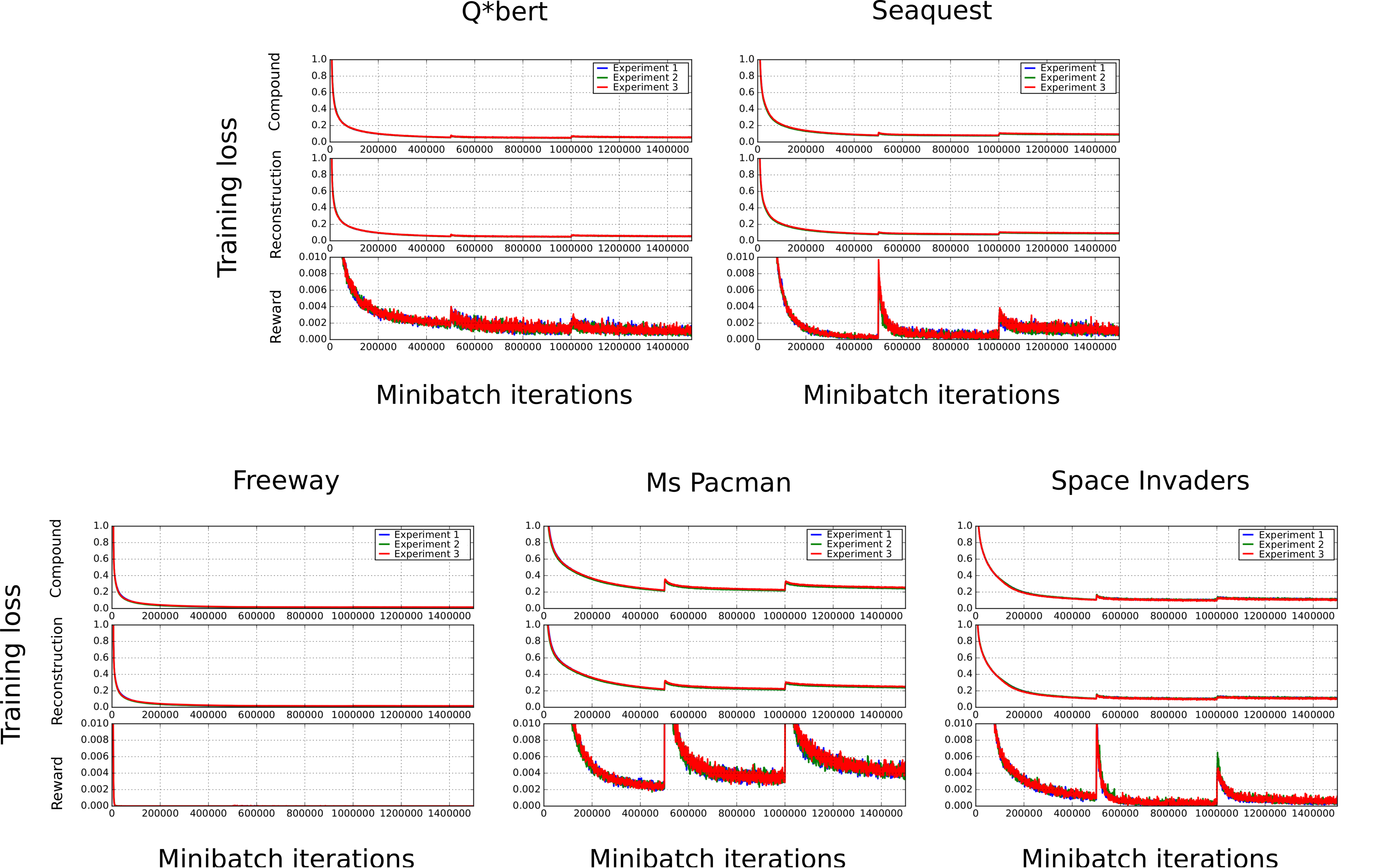}
\caption{Effect of curriculum learning on five different Atari games. Each panel corresponds to a different game, individual panels are structured in the same way as are those in Figure~\ref{fig:eff_grad_clip}}
\label{fig:eff_curr}
\end{center}
\end{figure*}

In our final experiments with curriculum learning, the networks were trained for $1,500,000$ minibatch iterations in total but the look ahead parameter $K$ was gradually increased every $500,000$ iterations from 1 to 3 to 5. The networks were hence initially trained on one-step ahead prediction only and later on fine-tuned on further-step ahead prediction. Figure~\ref{fig:eff_curr} shows how the training objective evolves over iterations. The characteristic "bumps" in the training objective every $500,000$ iterations as training evolves demonstrate improvements in long-term predictions in all games except Freeway where the training objective assumed already very low values within the first $500,000$ iterations and might have been therefore insensitive to further fine-tuning by curriculum learning.

\subsection{Effect of Random Seeds}
\label{eff_seed_exp}

\begin{figure*}[h]
\begin{center}
\includegraphics[width=1\linewidth]{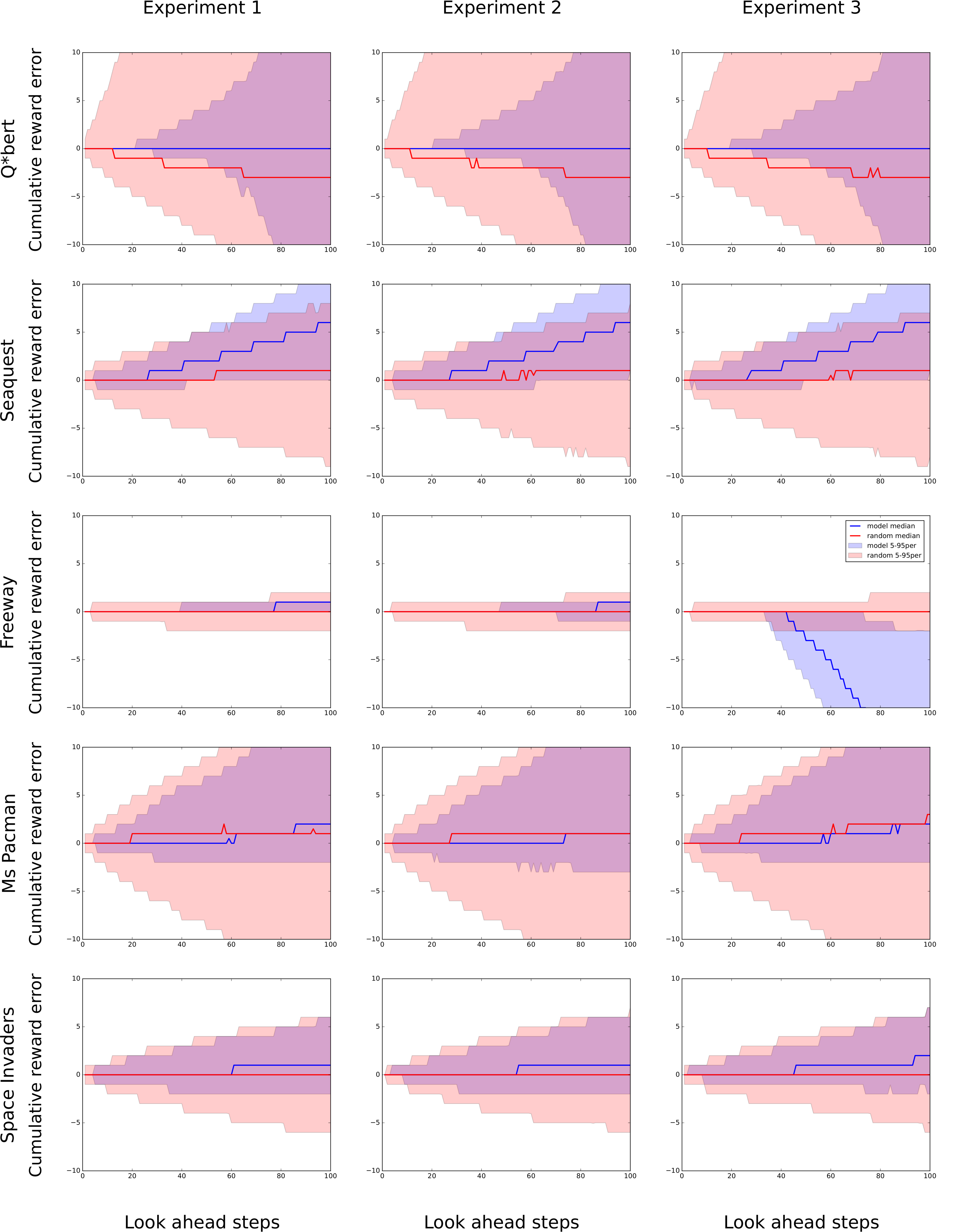}
\caption{Effect of different initial random seeds on cumulative reward error. The plots show how the cumulative reward error evolves over look ahead steps in terms of the median and the 5~to 95~percentiles for our network model (blue) as well as the baseline model (red) in each experiment. Each row refers to a different game, each column refers to a different experiment per game initialized with a different random seed. The first column of this figure is presented in Figure~\ref{fig:cumul_err} of the main paper explaining the results in more detail by additionally illustrating empirical distributions over the cumulative reward error for some look ahead steps.}
\label{fig:eff_sim}
\end{center}
\end{figure*}

We conducted three different experiments per game with different initial random seeds. The effect of different initial random seeds on the cumulative reward error is summarized in Figure~\ref{fig:eff_sim} which reports how the median and the 5~to 95~percentiles of the cumulative reward error evolve over look ahead steps in the different experiments per game. Note that the results of the first column in Figure~\ref{fig:eff_sim} are shown in Figure~\ref{fig:cumul_err} from the main paper together with a more detailed analysis depicting empirical cumulative reward error distributions for some look ahead steps. The random initial seed does not seem to have a significant impact on the cumulative reward prediction except for Freeway where the network in the third experiment starts to considerably overestimate cumulative rewards at around 30 to 40 look ahead steps.

\begin{figure*}[h]
\begin{center}
\includegraphics[width=0.9\linewidth]{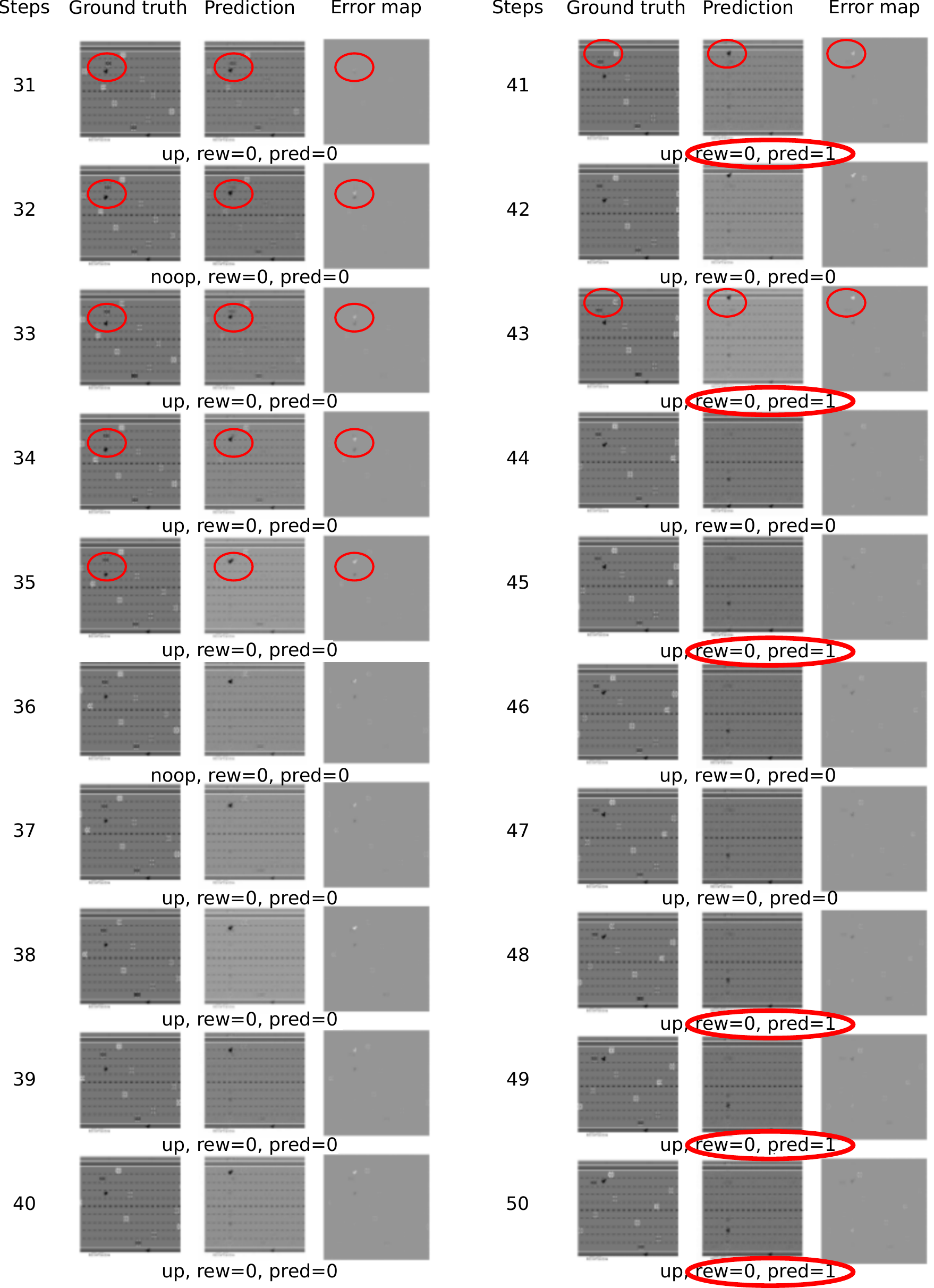}
\caption{Example predictions in Freeway over 20 steps. The figure is similar in nature to Figure~\ref{fig:trajecs_s} from the main paper with the only difference that predictions are depicted from time step 31 onwards.}
\label{fig:trajecs_f}
\end{center}
\end{figure*}

In order to investigate this reward overestimation in Freeway further, we analyse visualizations of joint video frame and reward prediction for this particular seed (similar in style to Figure~\ref{fig:trajecs_s} from Section~\ref{traj_sea} in the main paper). The results are shown in Figure~\ref{fig:trajecs_f} where a peculiar situation occurs after 31 predicted look ahead steps. In Freeway, the agent's job is to cross a busy road from the bottom to the top without bumping into a car in order to receive reward. If the agent bumps into a car, the agent is propelled downwards further away from the reward-yielding top. This propelled downwards movement happens even when the agent tries to move upwards. Exactly that kind of situation is depicted at the beginning of Figure~\ref{fig:trajecs_f} and occurs for this particular prediction after 31 steps. Our predictive model is however not able to correctly predict the aforementioned downwards movement caused by the agent hitting the car, which is highlighted in red throughout steps 31 to 35 documenting an increasing gap between ground truth and predicted agent position as the propelled downwards movement of the ground truth agent continues. In the course of further prediction, the network model assumes the agent to reach the reward-yielding top side of the road way too early which results in a sequence of erroneous positive reward predictions throughout steps 41 to 50, and as a side effect seemingly that the predictive model loses track of other objects in the scene. Concluding, this finding may serve as a possible explanation for cumulative reward overestimation for that particular experiment in Freeway. 

\subsection{Loss on Test Set}
\label{test_loss}

\begin{figure*}[h]
\begin{center}
\includegraphics[width=0.9\linewidth]{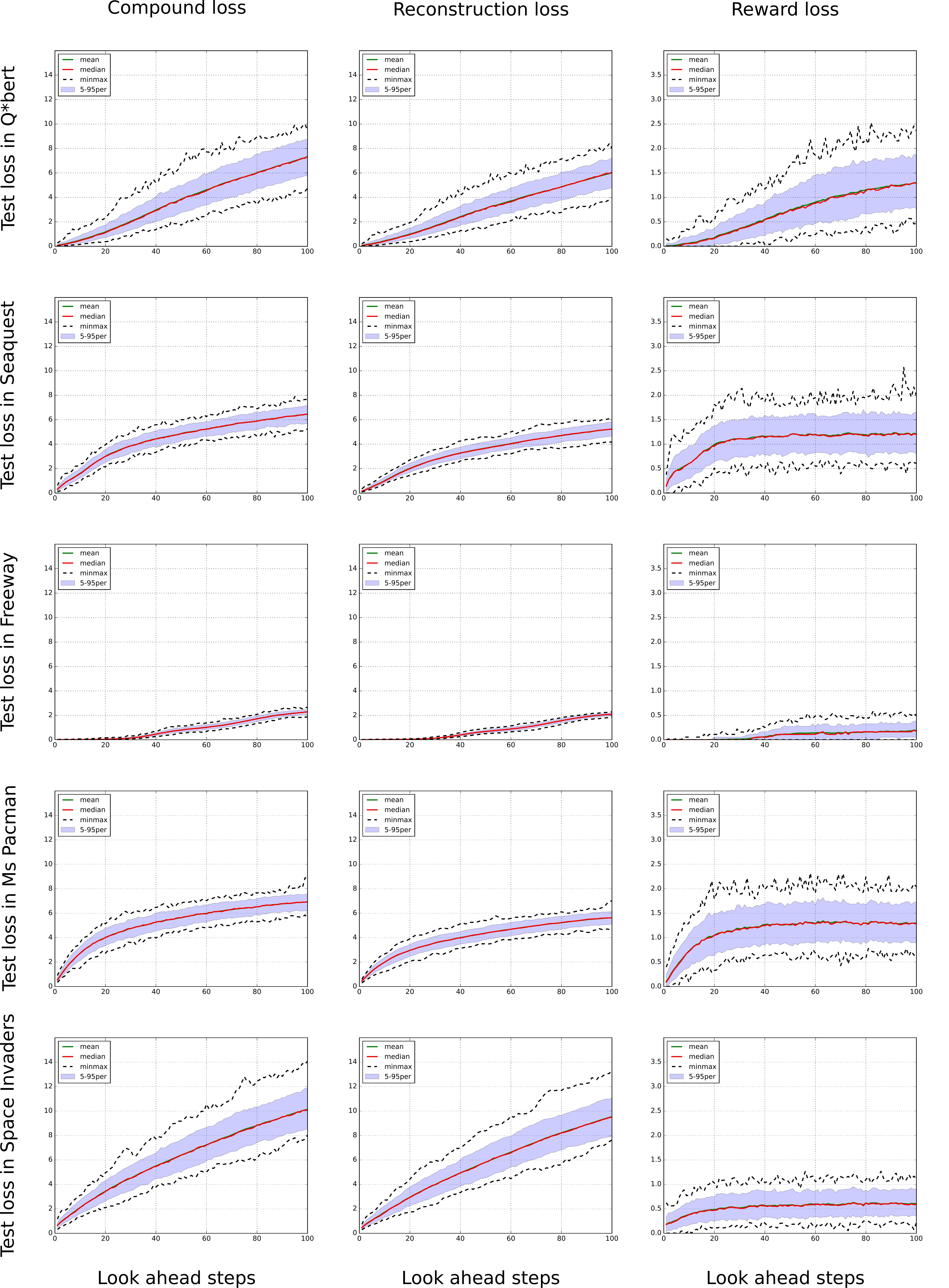}
\caption{Loss on test set over look ahead steps. Each row reports the loss on the test set over 100 look ahead steps for a different game. The first column illustrates the compound loss consisting of the video frame reconstruction loss (second column) and the reward prediction loss (third column). The loss on the test set is computed according to \cite{Oh2015} similar to the training loss for learning network parameters, however with a different look ahead parameter $K=100$ and a different minibatch size $I=50$, and without averaging over look ahead steps since we aim to plot the test loss as a function of look ahead steps. For each game, the test loss is computed for 300 minibatches resulting in an empirical distribution with 300 loss values per look ahead step. The figure shows the mean (in green), the median (in red), the 5~to 95~percentiles (in shaded blue) as well as minimum and maximum elements (in black dashed lines) of these empirical distributions.}
\label{fig:test_loss}
\end{center}
\end{figure*}

In the main paper, our analysis focuses on evaluating how well our model serves the purpose of cumulative reward prediction. Here, we evaluate network performance in terms of both the video frame reconstruction loss as well as the reward prediction loss on the test set following the analysis conducted in \cite{Oh2015}. For each game, we sample 300 minibatches of size $I=50$ from the underlying test set and compute the test loss over $K=100$ look ahead steps with the formula presented in the main paper in Section~\ref{train} used for learning network parameters, but without averaging over look ahead steps because we aim to illustrate the test loss as a function of look ahead steps---statistics of this analysis are plotted in Figure~\ref{fig:test_loss}. 

Best overall test loss is achieved in Freeway and for initial look ahead steps (up to roughly between 40 and 60 steps) in Q*bert, which is in accordance with results for cumulative reward prediction from the main paper. Also in line with results from the main paper is the finding that the reward loss on the test set is worse in Seaquest, Ms Pacman and Space Invaders when compared to Q*bert (up to approximately 40 steps) and Freeway. Worst video frame reconstruction loss is observed for Space Invaders in compliance with \cite{Oh2015} where the authors report that there are objects in the scene moving at a period of 9 time steps which is hard to predict by a network only taking the last 4 frames from the last 4 steps as input for future predictions. At first sight, it might seem a bit surprising that the reward prediction loss in Space Invaders is significantly lower than in Seaquest and Ms Pacman for long-term ahead prediction despite the higher frame reconstruction loss in Space Invaders. A possible explanation for this paradox might be the frequency at which rewards are collected---this frequency is significantly higher in Seaquest and Ms Pacman than in Space Invaders. A reward prediction model with bias zero---as indicated in the paper---might therefore err less often when rewards are collected at a lower frequency hence yielding lower reward reconstruction loss.

\subsection{Comparison to Training with Separate Objectives}
\label{comparison}

We could furthermore ascertain that using a joint objective for video frame reconstruction and reward prediction is beneficial. To this end, we conducted experiments with two baseline approaches using separate objectives for video frame and reward prediction. The first baseline uses the same architecture from the main paper but with a decoupled training objective. This means that the reward prediction part from the network shown in Figure~\ref{fig:model} is trained using the hidden representation from the video frame prediction part as input. Importantly, gradient updates with respect to reward prediction do no longer affect the video frame prediction parameters as they would when training is conducted with a joint objective.

The second baseline uses a decoupled network architecture with two different convolutional networks for video frame and reward prediction that are trained separately from one another. The video frame prediction network of the decoupled architecture is modelled according to \cite{Oh2015} depicted in Figure~\ref{fig:model} but without the reward prediction part highlighted in red, whereas the reward prediction network is modelled similar to Figure~\ref{fig:model} with an encoding stage, an action-conditioned transformation stage and a reward prediction layer, but without a video frame decoding stage. 

\begin{figure*}[h]
\begin{center}
\includegraphics[width=1\linewidth]{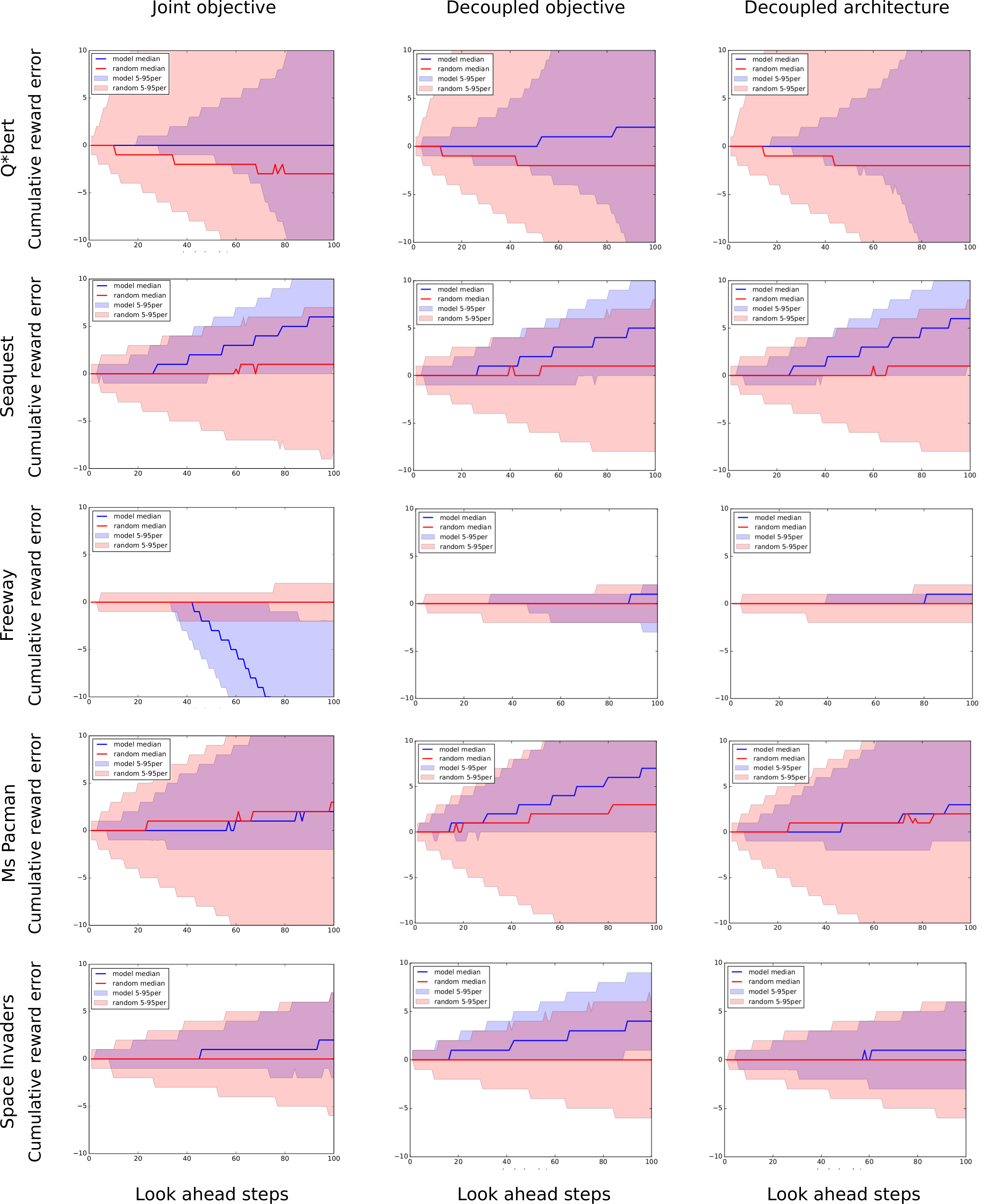}
\caption{Effect of different training methods on cumulative reward error. The plots show how the cumulative reward error evolves over look ahead steps in terms of the median and the 5~to 95~percentiles for our network models (blue) as well as the random prediction baseline model (red). Each row refers to a different game. Each column refers to a different training method. The first column refers to training with a joint objective for video frame and reward prediction as outlined in the main paper and presented in the third column of Figure~\ref{fig:eff_sim}---note that this contains the outlier for Freeway. The second column refers to training with a decoupled objective where the reward prediction part of the network in Figure~\ref{fig:model} is trained separately using the hidden state of the video frame prediction model as input. The third column refers to training with a decoupled architecture with two separate convolutional networks, one for video frame reconstruction and one for reward prediction. The overall result is that training with a decoupled objective works worse compared to using a joint training objective or a decoupled architecture. The joint objective works equally well as the decoupled architecture but requires significantly less parameters due to a shared network architecture.}
\label{fig:eff_train}
\end{center}
\end{figure*}

The results are shown in Figure~\ref{fig:eff_train}. Overall, training with a decoupled objective is significantly worse compared to training with a joint objective and training with a decoupled architecture in all games except Seaquest (note that experiments were conducted with one seed only containing the outlier from Freeway in Figure~\ref{fig:eff_sim}). The joint objective works equally well as the decoupled architecture with the added benefit that less parameters are required due to a shared network architecture. A shared network architecture with a single latent representation encoding both dynamics and the instantaneous reward signal might also be advantageous for future work on computationally efficient planning in lower-dimensional latent spaces. 

\end{document}